\documentclass{article}
\usepackage{spconf,amsmath,graphicx}
\usepackage{amssymb}
\usepackage[utf8]{inputenc}
\usepackage{newunicodechar}
\newunicodechar{ℝ}{\mathbb{R}}
\usepackage{booktabs,chemformula}

\usepackage{xcolor}

\usepackage{enumitem}
\setlist{nosep, leftmargin=14pt}

\usepackage{mwe} 

\title{Enhanced Masked Image Modeling for Analysis of Dental Panoramic Radiographs}

\name{Amani Almalki and Longin Jan Latecki
\vspace{-5mm}}
\address{Department of Computer and Information Sciences, Temple University, Philadelphia, USA\\
{\tt 
\{amani.almalki,latecki\}@temple.edu}}

\begin{document}
%
\maketitle
\begin{abstract}
\vspace{-1mm}
The computer-assisted radiologic informative report has received increasing research attention to facilitate diagnosis and treatment planning for dental care providers. However, manual interpretation of dental images is limited, expensive, and time-consuming. Another barrier in dental imaging is the limited number of available images for training, which is a challenge in the era of deep learning. This study proposes a novel self-distillation (SD) enhanced self-supervised learning on top of the masked image modeling (SimMIM) Transformer, called SD-SimMIM, to improve the outcome with a limited number of dental radiographs. In addition to the prediction loss on masked patches, SD-SimMIM computes the self-distillation loss on the visible patches. We apply SD-SimMIM on dental panoramic X-rays for teeth numbering, detection of dental restorations and orthodontic appliances, and instance segmentation tasks. Our results show that SD-SimMIM outperforms other self-supervised learning methods. Furthermore, we augment and improve the annotation of an existing dataset of panoramic X-rays.

\end{abstract}


\vspace{-3mm}
\section{Introduction}
\label{sec:intro}
\vspace{-3mm}

The computer-assisted decisions are essential in dental practice to help dentists diagnose and plan for treatments. Dental imaging is a valuable tool that facilitates diagnosis and treatment plans, which is impossible through clinical examination and patient history only \cite{white2001parameters}. A Dental X-ray is a two-dimensional radiograph that captures the patient's entire mouth from ear to ear in a single image, including the upper and lower jaws and surrounding alveolar bone \cite{rushton1996use}. 

In dentistry, many teeth numbering systems provide a specific code for each tooth. Specifically, in this study, we utilize The Federation Dentiaure International numbering system (FDI), which is internationally known among dental care providers. It is a two-digit code where the first digit is given for each quadrant from 1 to 4 for permanent adult teeth. And the second digit is assigned for each tooth number based on its location in the jaw, starting from the middle front teeth (number 1) and moving back up to the third molar (number 8) \cite{smith1976numbering}. 

Furthermore, dental restorations are used to restore the tooth's missing structure resulting from caries or trauma with full or partial coverage. Moreover, root canal fillings are utilized to fill the space of the root portion inside the tooth structure because of decay or other damage. In addition, orthodontic appliances apply force onto the teeth to be moved into the correct position; such appliances include but are not limited to bands, brackets, and retainers. The restorative materials and orthodontic appliances appear radiopaque in the X-rays and can be identified by dental practitioners \cite{molander1996panoramic}.

Deep learning models are successful when trained with a large amount of data, however, a very limited number of dental radiographs is available for training. To mitigate this problem, we propose a new self-distillation and self-supervised learning combination for training a Swin Transformer \cite{liu2021swin} for dental panoramic X-rays analysis. 

Recently, self-supervised learning methods with masked image modeling (MIM) such as SimMIM \cite{xie2022simmim}, MAE \cite{he2022masked} and UM-MAE \cite{li2022uniform} are shown to be effective in pre-training deep learning models, like Transformers \cite{liu2021swin, dosovitskiy2020vit}. However, only SimMIM and UM-MAE are applicable to Swin Transformer. Generally, the idea of MIM methods is to mask some patches before they are fed into the Swin encoder and predict the original patches to gain more understanding of the images. However, the patches' location is important in dental panoramic X-rays for a predictable outcome. SimMIM maintains the patches location known to both the encoder and decoder, while UM-MAE drops the location information unknown to the encoder, which may induce inaccuracy. Therefore, SimMIM pre-training is selected in this study.

Inspired by \cite{xie2022simmim, he2022masked, luo2022self}, we hypothesize that the Swin encoder can be improved by transferring knowledge obtained by decoded visible patches to their encoded peers through self-distillation. We believe that the visible patches in the decoder contain more knowledge than the ones in the encoder. Moreover, similar to \cite{xie2022simmim, he2022masked} and unlike \cite{luo2022self}, we found out that predicting the masked area only outperforms predicting all image pixels.

The proposed SD-SimMIM is trained on the same dataset as the downstream tasks, excluding the test dataset. We apply SD-SimMIM on dental panoramic X-rays for teeth numbering, detection of dental restorations and orthodontic appliances, and instance segmentation tasks. It is shown that SD-SimMIM performs better than other self-supervised learning methods.

Although previous studies investigated teeth numbering \cite{silva2020study} and segmentation of dental restorations \cite{almalki2023self}, there is no comprehensive dataset that simultaneously studied orthodontics appliance segmentation. We believe that the inclusion of segmentation of orthodontics appliances increases the complexity of the computer vision problem because of class quantity and class imbalance. Therefore, we augment the existing dataset introduced in \cite{silva2020study} under dental expert supervision. We further expand the dataset by developing annotations for orthodontics appliances, including bands, brackets, and retainers. The labeling process led to a unique high-quality augmented dataset. Our data will be available, upon request, under the name {\bf Dent}al an{\bf alysis} (Dentalysis) annotations.
Our main contributions are twofold: 
\begin{itemize}
\item We introduce SD-SimMIM, a self-distillation enhanced SimMIM. It aims to boost the feature representation on top of SimMIM to alleviate the demands on large data for dental panoramic radiographs, and further help downstream tasks.
\item The augmented dataset increases performance, while added labeling of orthodontics appliances extends the horizon of possible dental applications.
\end{itemize}

\vspace{-3mm}
\graphicspath{{images/}}
\section{Methods}
\label{sec:methods}
\vspace{-3mm}

\begin{figure}
\begin{center}
\includegraphics[width=1\linewidth]{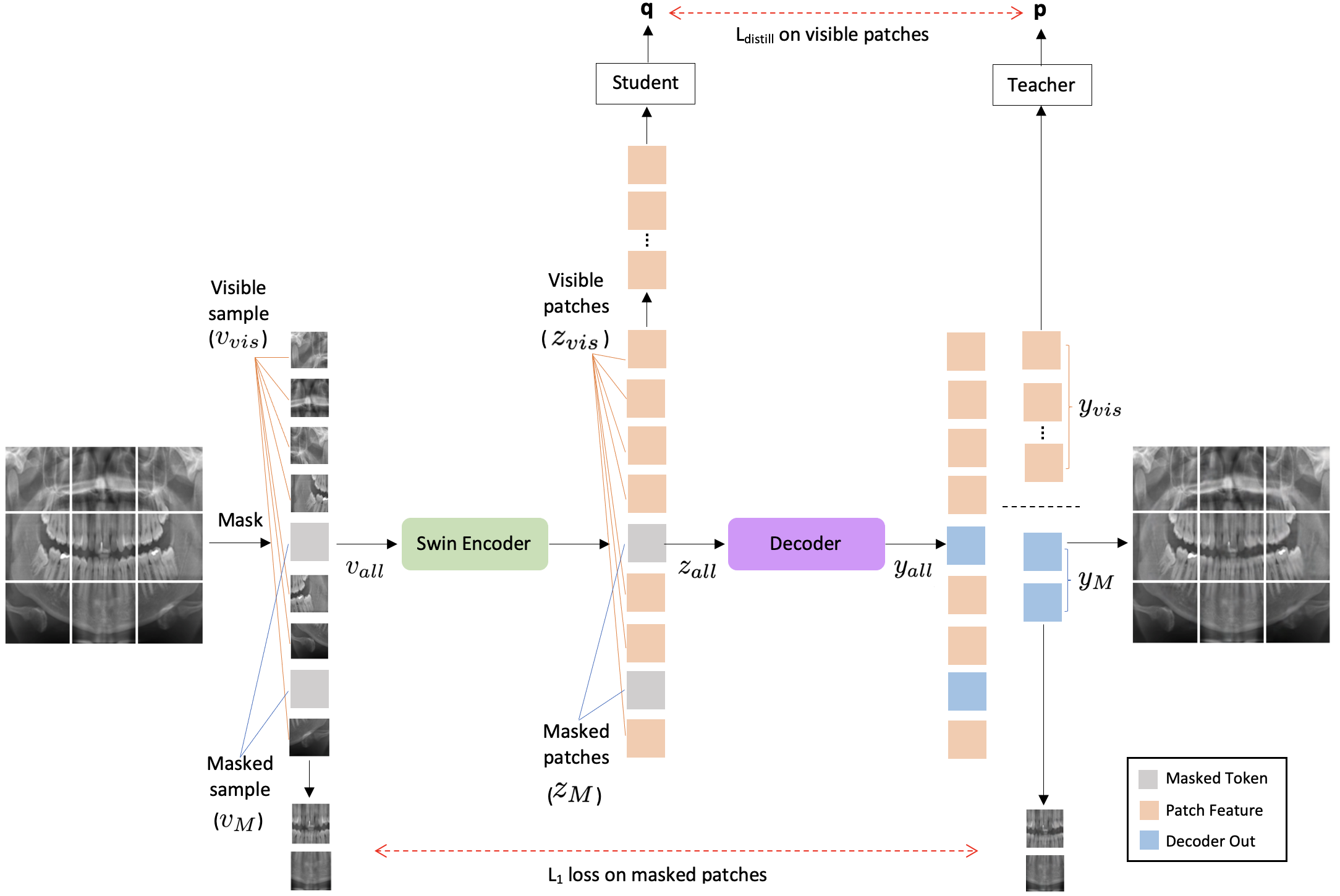} 
\vspace{-11mm}
\end{center}
   \caption{{\bf Our SD-SimMIM framework.} Alongside the original SimMIM, we benefit from decoded visible patches (as the teacher) and transfer knowledge to their peers after encoding. (Best viewed in color)}
\label{fig:pipeline}
\end{figure}

Fig.~\ref{fig:pipeline} illustrates our SD-SimMIM framework. It includes two modules, masked image modeling (MIM) and visible image modeling (VIM). MIM generates self-supervised learning on unlabeled data by masking some image patches, while VIM imposes self-distillation constraints on visible patches for better and more powerful encoder learning. Hence, VIM enhances the original SimMIM, particularly for dental panoramic radiographs.  

\subsection{SimMIM}
\label{ssec:sim}
\vspace{-3mm}

SimMIM framework includes four components: patchifying and masking, encoder, decoder, and prediction target.

{\bf Patchifying and masking} designs how to select the area to mask, and how to implement masking of the selected area. The Patchifying first divides the input image $x$ into $N$ patches. Then, it flattens each patch to a token (a one-dimensional vector of visual features) with length $D$. Hence, the formulation of the representation of all patches is $v_{all} \subseteq {\mathbb{R}}^{N\times{D}}$. Next, the masking randomly divides the patches into two sets with respect to a masking ratio $M$, more precisely $v_{all} \subseteq {\mathbb{R}}^{N\times{D}}\rightarrow {v_{vis}} \subseteq {\mathbb{R}}^{N'\times{D}},{v_{M}} \subseteq {\mathbb{R}}^{\tilde{N'}\times{D}}$ where $N'=N \times (1-M), \tilde{N'} =N \times M$. $v_{all}$ will be the input to the encoder and $v_{M}$ are the labels.

{\bf Encoder} takes $v_{all}$ as input, and extracts latent feature from visible patches. First, it maps $D$ dimensions of tokens to $D'$ with a linear projection, and then these patch tokens are processed via Swin Transformer blocks to get latent representation vectors of patches ${z_{vis}} \subseteq {\mathbb{R}}^{N'\times{D'}}$ and masked tokens ${z_{M}} \subseteq{\mathbb{R}}^{\tilde{N'}\times{D'}}$.

{\bf Decoder} takes ${z_{all}} \subseteq{\mathbb{R}}^{{N}\times{D'}}$ as input, and learns low-level representation from visible patches for image reconstruction. Hence, the decoder output ${y_{all}} \subseteq{\mathbb{R}}^{{N}\times{D'}}$ will divided into $y_{vis}$ and $y_{M}$, as visible and masked tokens, respectively.

{\bf Prediction target} defines the form of original signals to predict. First, we consider the original masked tokens after normalizing $Y_{M} = Norm(v_{M})$ as our prediction target. The decoder applies a linear layer to align $y_{M}$ and $Y$, i.e. $y_{M} \rightarrow y'_{M}$. The $L_{1}$ loss is computed between the predicted masked tokens $y'_{M}$ and the original masked tokens after normalization $Y_{M}$ as described in Eq.~\ref{eqn:eq1}.

\begin{equation}
\label{eqn:eq1}
L_{1} = {\ell_{1}(y'_{M},Y_{M})}, \quad {y'_{M},Y_{M}} \subseteq {\mathbb{R}}^{\tilde{N'}\times{D}}
\end{equation}

\subsection{Self-distillation}
\label{ssec:SD}
\vspace{-3mm}

Knowledge Distillation is the process of transferring knowledge from a large model to a smaller one \cite{hinton2015distilling}. Previous studies apply it to the vectors at various depths within the same network, either a convolutional neural network (CNN) \cite{zhang2019your} or a Vision Transformer (ViT)\cite{luo2022self}. Hence, knowledge is distilled from deep layers to shallow layers, augmenting the feature representation of shallow layers. Considering the imbalance of knowledge, we found that this is exactly how knowledge in the visible tokens can be transferred from the decoder to the encoder through this distillation paradigm. Particularly, there are two types of latent representation vectors for visible tokens in SimMIM, i.e. $z_{vis}$ outputted from the encoder and $y_{vis}$ from the decoder. We treat $z_{vis}$ as shallow features and $y_{vis}$ as deeper features in the self-distillation framework \cite{zhang2019your}. We use a 3-layer MLP over these two vectors, resulting in probability distributions over $K$ dimensional feature denoted by $q$ and $p$, respectively. Each of them is normalized with a $Softmax$ over the feature dimension. Thus, we learn to match these distributions by minimizing the cross-entropy loss as shown in Eq.~(\ref{eqn:eq2}).

\begin{equation}
\label{eqn:eq2}
    \begin{split}
        & q = MLP(z_{vis}),\quad p = MLP(y_{vis}) \\
        & q' = Softmax(q),\quad p' = Softmax(p) \\
        & L_{distill} = -p'log(q')
    \end{split}
\end{equation}

The total loss is formulated as shown in Eq.(\ref{eqn:eq3}).

\begin{equation}
\label{eqn:eq3}
L = \alpha L_{1} + (1-\alpha) L_{distill} 
\end{equation}
where $\alpha$ is the empirically defined scaling factor (in this study, $\alpha$ is equal to 0.2).

\vspace{-3mm}
\section{Experiments}
\label{sec:experiments}
\vspace{-3mm}

\subsection{Dataset}
\label{ssec:dataset}
\vspace{-3mm}

\begin{figure}
\begin{center}
\includegraphics[width=1\linewidth]{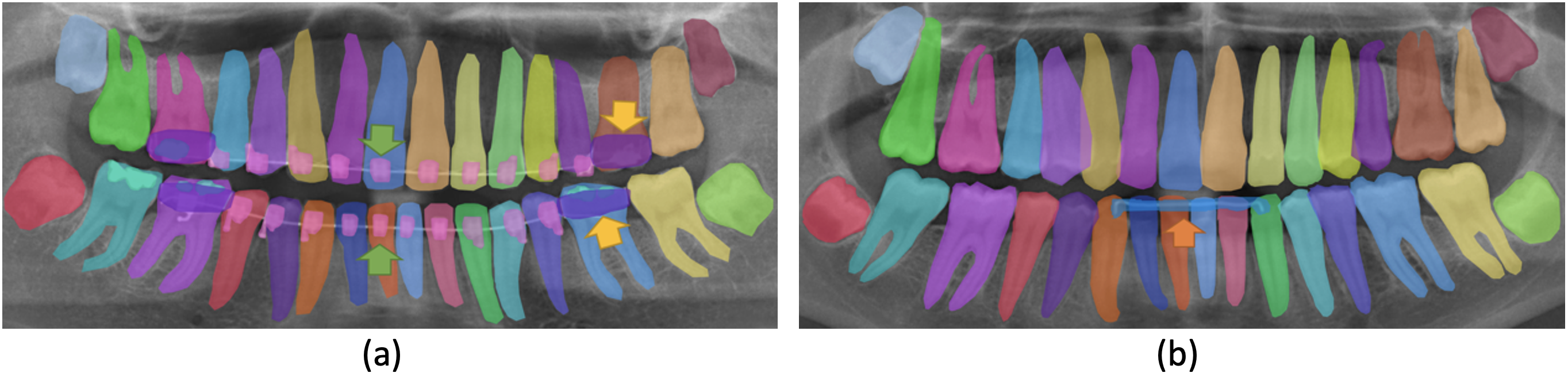}
\vspace{-11mm}
\end{center}
  \caption{Samples of segmentation of orthodontics appliances, a) shows examples of bands (yellow arrows) and brackets (green arrows), and b) a retainer (orange arrow). (Best viewed in color)}
\label{fig:samples}
\end{figure}

Detection, Numbering, and Segmentation (DNS) \cite{silva2020study} is a dental panoramic X-rays dataset consisting of 543 annotated images with ground truth segmentation labels, including numbering information based on the FDI teeth numbering system. Each image size is 1991x1127 pixels. The dataset annotations from \cite{almalki2023self} do not contain any segmentation of orthodontic appliances. Therefore, we contribute to expanding the dataset by developing segmentation for orthodontic appliances and introducing three more classes, namely bands, brackets, and retainers. This process was under a supervision of a dentist using the COCO-Annotator tool \cite{cocoannotator}. We attended weekly meetings where related issues and questions were discussed. In the end, the annotations were reviewed to assure quality and avoid systematic and random errors. Fig.~\ref{fig:samples} presents samples of segmentation of orthodontics appliances. We believe this is the most inclusive dataset for segmenting teeth, dental restorations, and orthodontic appliances in dental panoramic radiographs. Our data will be available upon request, namely Dentalysis annotations.

\subsection{Evaluation metric}
\vspace{-3mm}

For all our experiments, we split the data into five folds, each containing about 20\% of the images. One fold is fixed as the test set (111 images), and the other four folds (108 images each) compose the training and validation datasets in a cross-validation manner. This process is repeated five times. The evaluation metric we adopt is the Average Precision for object detection and instance segmentation models.

\subsection{Implementation details}
\vspace{-3mm}

Our experiments are implemented based on the PyTorch \cite{paszke2019pytorch} framework and trained with NVIDIA Tesla Volta V100 GPUs. In all experiments, the batch size equals the total number of training samples, which is 432. The input images are all resized to 800×600 pixels. We utilize the AdamW \cite{loshchilov2017decoupled} optimizer in all experiments.

{\bf Data augmentation.} We apply noise addition and horizontal flipping, which turns left teeth numbers into right teeth numbers and vice-versa.

{\bf SD-SimMIM pre-training.} We follow a similar protocol to SimMIM \cite{xie2022simmim} to train our SD-SimMIM. We use Swin-B \cite{liu2021swin} as the encoder and a lightweight decoder with a linear projection. The base learning rate is set to 8e-4, weight decay is 0.05, $\beta1$ = 0.9, $\beta2$ = 0.999, with a cosine learning rate scheduler. We use a random MIM with a patch size of 16×16 and a mask ratio of 20\%. We apply the L2-normalization bottleneck \cite{caron2021emerging} (dimension 256 for the bottleneck and $K$ dimensions equals 4096) as the projection head in self-distillation. This model was pre-trained for 100 epochs with a warm-up for 10 epochs. The target image size is 800×600.

{\bf Task fine-tuning.} We utilize single-scale training. The initial learning rate is 0.0001, and the weight decay is 0.05.

\subsection{Quantitative results}
\label{sec:quan}
\vspace{-3mm}

Table~\ref{table:results} shows the results of different methods on the dataset for teeth numbering, detection of dental restorations, and instance segmentation only. As a baseline (the first row, called Supervised), Swin-B \cite{liu2021swin} is trained using the dataset without self pre-training to demonstrate the improvement obtained by self-supervised learning. The original Swin-B was trained on the Image Net dataset with 1000 classes denoted as (IN-1K).
The CNN-based network, PANet \cite{silva2020study}, reports a result that is worse than Swin-B. This can be explained as the difference in the network capacity, where ResNet-50 is used as the backbone in PANet. 
As a comparison, the way SimMIM uses image reconstruction is obviously more suitable than UM-MAE for dental images. The reason may be attributed to the fact that the location of the patches is essential in dental radiographs for a predictable outcome. SimMIM maintains the location of the patches known to both the encoder and decoder, while UM-MAE drops the location information, which may induce inaccuracy. 
The proposed SD-SimMIM shows steady improvements over SimMIM and yields the best performance.
Hence transferring decoder information to the encoder with self-distillation improves  the outcomes of self-learning. 
We also observe that similar to \cite{xie2022simmim, he2022masked} and unlike \cite{luo2022self}, our results show that predicting the masked area only outperforms predicting all image pixels for both SimMIM and our SD-SimMIM.

\begin{table}[h]
\centering
\caption{Results of teeth numbering, detection of dental restorations, and instance segmentation only. * denotes $L_{1}$ loss is computed on the whole image. ${AP}^{box}$ and ${AP}^{mask}$ indicate Average Precision for object detection and instance segmentation, respectively.}
{\footnotesize 
\begin{tabular}{@{}ccccc@{}}
\toprule
Initialization & Backbone  & Pre-train Data & ${AP}^{box}$ & ${AP}^{mask}$ \\
\midrule
\ch Supervised & Swin-B  & IN-1K w/ Labels  & 80.3  & 79.2 \\
\midrule
\ch PANet \cite{silva2020study} & ResNet-50 & IN-1K w/ Labels   & 76.8  & 75.1   \\ 
\ch UM-MAE  \cite{almalki2023self}   & Swin-B    & IN-1K  & 88.3  & 85.7   \\
\ch SimMIM*  & Swin-B  & IN-1K  & 89.9  & 88.5 \\
\ch SimMIM \cite{almalki2023self} & Swin-B  & IN-1K  & 90.4  & 88.9 \\
\midrule
\ch SD-SimMIM*  & Swin-B  & IN-1K  &  90.7  &  89.6   \\
\ch SD-SimMIM & Swin-B & IN-1K &  {\bf 92.4} &  {\bf 90.2} \\ 
\bottomrule
\end{tabular}
}
\label{table:results}
\end{table}

Table~\ref{table:resultsOrth} shows results after including the annotations of orthodontics appliances. The proposed SD-SimMIM method achieves the highest performance of 92.7\% and 90.8\% on detecting teeth, dental restorations and orthodontics appliances, and instance segmentation, respectively. Again it is worth noting that the best performance is gained when computing the loss on the masked areas only. 

\begin{table}[h]
  \centering
  \caption{Results after including orthodontics appliances. * denotes $L_{1}$ loss is computed on the whole image.}
  {\footnotesize 
  \begin{tabular}{@{}ccccc@{}}
    \toprule
Initialization & Backbone  & Pre-train Data & ${AP}^{box}$ & ${AP}^{mask}$ \\
  \midrule
\ch Supervised & Swin-B  & IN-1K w/ Labels  & 81.9  & 80.1 \\
\midrule
\ch SimMIM*   & Swin-B    & IN-1K   &  90.3  &  88.8  \\ 
\ch SimMIM \cite{almalki2023self} & Swin-B  & IN-1K  & 90.8  & 89.4 \\
\midrule
\ch SD-SimMIM*     & Swin-B    & IN-1K    &  91.2  &  90.0   \\
\ch SD-SimMIM & Swin-B & IN-1K &  {\bf 92.7} &  {\bf 90.8}
\\ 
    \bottomrule
  \end{tabular}
  }
  \label{table:resultsOrth}
\end{table}

\vspace{-3mm}
\subsection{Qualitative results}
\vspace{-3mm}

\begin{figure}
\begin{center}
\includegraphics[width=1\linewidth]{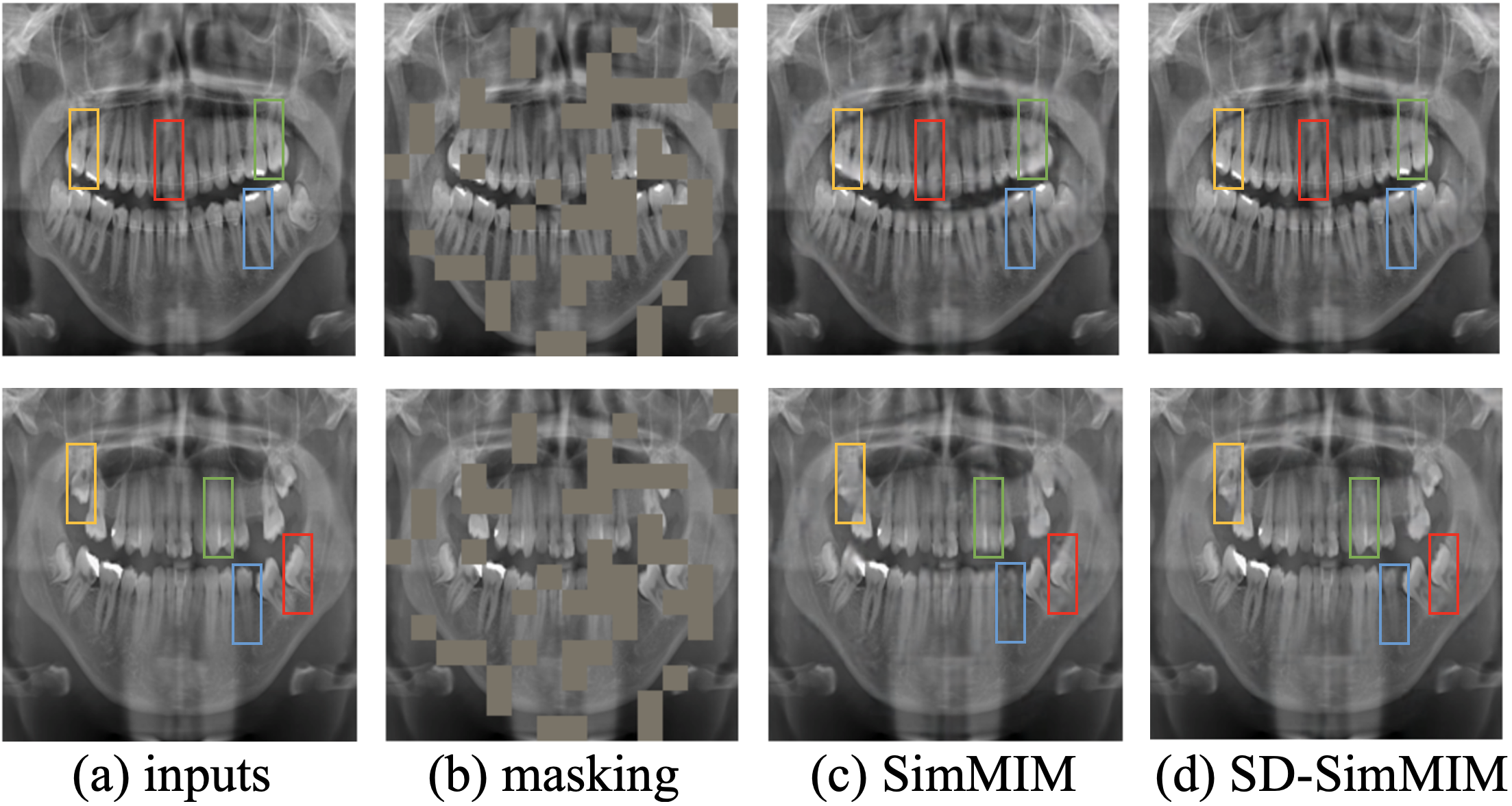}
\vspace{-11mm}
\end{center}
   \caption{Images reconstructed by SimMIM and SD-SimMIM. SD-SimMIM shows a clearly better reconstruction than SimMIM. The color boxes highlight their details. (Best viewed in color)}
\label{fig:masking}
\end{figure}

To illustrate the effectiveness of adding self-distillation to simMIM, we provide some visualization examples. Firstly, we are curious about the results of image reconstruction. Fig.~\ref{fig:masking} presents two reconstruction examples using our SD-SimMIM. As shown, SD-SimMIM obtains a slightly better reconstruction than SimMIM. It proves that self-distillation reinforces the learning capability of the SimMIM encoder.

\begin{figure}
\begin{center}
\includegraphics[width=1\linewidth]{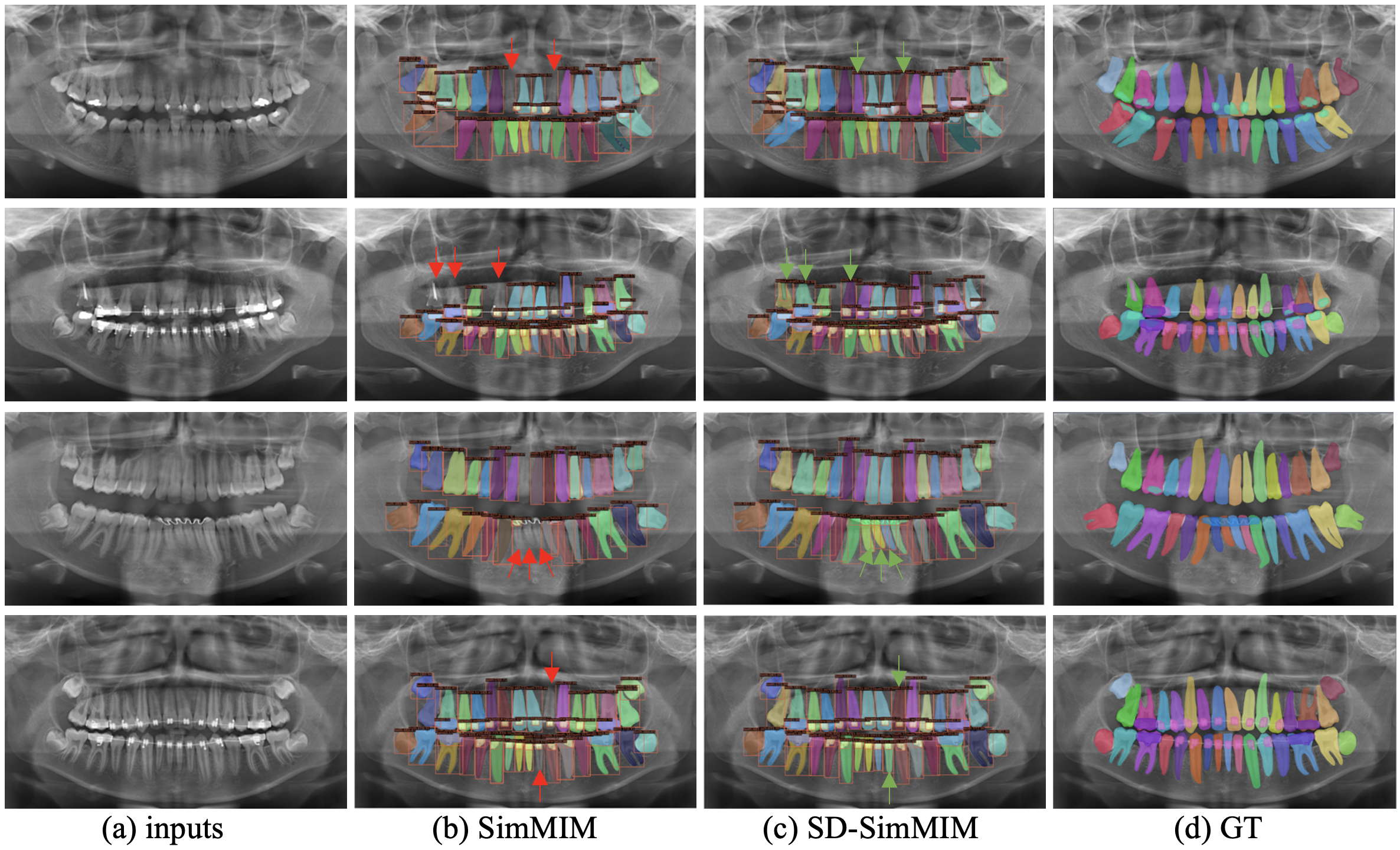} 
\vspace{-11mm}
\end{center}
   \caption{Qualitative results of detection and instance segmentation. Note that teeth detection and instance segmentation are missing (red arrows) when Swin Transformer is pre-trained with SimMIM compared to the ones produced by Swin Transformer pre-trained with SD-SimMIM architecture (green arrows). (Best viewed in color.)}
\label{fig:DetSeg}
\end{figure}

Secondly, Fig.~\ref{fig:DetSeg} displays four different qualitative samples of improved performance when the Swin Transformer is pre-trained with SD-SimMIM for teeth numbering, detecting dental restorations, orthodontic appliances, and instance segmentation. Those improvements in detection and segmentation agree with the quantitative results in Section~\ref{sec:quan}.

\section{Conclusions}
\label{sec:conclusions}

We propose SD-SimMIM, a novel self-distillation scheme that transfers knowledge from the decoder to the encoder to guide a more effective visual pre-training. The quantitative and qualitative results present the benefits of our SD-SimMIM, which is a promising tool for the analysis of dental radiographs. For future work, we will evaluate our SD-SimMIM on different downstream tasks such as detecting dental disease on dental bitewing radiographs.


\section{Compliance with ethical standards}
\label{sec:ethics}

This research study was conducted retrospectively using human subject data made available in open access by \cite{silva2020study}. Ethical approval was not required, as confirmed by the license attached to the open-access data.


\section{Acknowledgments}
\label{sec:acknowled gments}

We would like to express our deepest thanks to Dr. \mbox{Abdulrahman} Almalki, a dental expert from the University of Pennsylvania, for his valuable discussions related to dentistry. Also, we would like to thank Dr. Luciano Oliveira and his group from Ivisionlab at the Federal University of Bahia for sharing the dataset used in this study.

\clearpage

\bibliographystyle{IEEEbib}
\bibliography{strings,refs}

\end{document}